\documentclass[journal]{IEEEtran}

\usepackage[linesnumbered]{algorithm2e}
\usepackage{graphicx}
\usepackage{amsmath,amssymb,amsfonts,amsthm}
\usepackage{enumitem}
\usepackage{booktabs}
\usepackage{subcaption}
\usepackage{pgfplots}
\pgfplotsset{compat=newest}
\usepackage{multirow}
\usepackage{graphicx}
\usetikzlibrary{patterns}
\usepgfplotslibrary{statistics}
\usepgfplotslibrary{groupplots}
\pgfmathdeclarefunction{fpumod}{2}{%
    \pgfmathfloatdivide{#1}{#2}%
    \pgfmathfloatint{\pgfmathresult}%
    \pgfmathfloatmultiply{\pgfmathresult}{#2}%
    \pgfmathfloatsubtract{#1}{\pgfmathresult}%
    \pgfmathfloatifapproxequalrel{\pgfmathresult}{#2}{\def\pgfmathresult{5}}{}%
}
\newcommand{\spara}[1]{\smallskip\noindent{\bf #1}}
\newtheorem{thm}{Theorem}
\newtheorem{defn}[thm]{Definition}
\newtheorem{prob}{Problem}

\begin{document}

\title{Label-based Graph Augmentation with Metapath for Graph Anomaly Detection}

\author{Hwan~Kim,~
        Junghoon~Kim,~
        Byung~Suk~Lee,
        and~Sungsu~Lim,~\IEEEmembership{Member,~IEEE}

\thanks{H. Kim and S. Lim are with the Department of Computer Science and
Engineering, Chungnam National University, Daejeon 34134, Republic of
Korea (e-mail: hwan.kim@o.cnu.ac.kr, sungsu@cnu.ac.kr).}
\thanks{J. Kim is with the Department of Computer Science and Engineering, Ulsan National Institute of Science \& Technology, Ulsan 44191, Republic of Korea (e-mail: junghoon.kim@unist.ac.kr)}%
\thanks{B. S. Lee is with the Department of Computer Science, University of
Vermont, Burlington, Vermont 05405, USA (e-mail: bslee@uvm.edu).}
\thanks{\textit{(Corresponding author: Sungsu Lim.)}
}}

\maketitle

\begin{abstract}
Graph anomaly detection has attracted considerable attention from various domain ranging from network security to finance in recent years.
Due to the fact that labeling is very costly, existing methods are predominately developed in an unsupervised manner.
However, the detected anomalies may be found out uninteresting instances due to the absence of prior knowledge regarding the anomalies looking for.
This issue may be solved by using few labeled anomalies as prior knowledge. In real-world scenarios, we can easily obtain few labeled anomalies.
Efficiently leveraging labelled anomalies as prior knowledge is crucial for graph anomaly detection; however, this process remains challenging due to the inherently limited number of anomalies available.
To address the problem, we propose a novel approach that leverages metapath to embed actual connectivity patterns between anomalous and normal nodes.
To further efficiently exploit context information from metapath-based anomaly subgraph, we present a new framework, Metapath-based Graph Anomaly Detection (MGAD), incorporating GCN layers in both the dual-encoders and decoders to efficiently propagate context information between abnormal and normal nodes. 
Specifically, MGAD employs GNN-based graph autoencoder as its backbone network.
Moreover, dual encoders capture the complex interactions and metapath-based context information between labeled and unlabeled nodes both globally and locally.
Through a comprehensive set of experiments conducted on seven real-world networks, this paper demonstrates the superiority of the MGAD method compared to state-of-the-art techniques.
The code is available at https://github.com/missinghwan/MGAD.
\end{abstract}


\IEEEpeerreviewmaketitle

\section{INTRODUCTION} \label{sec:introduction}
\IEEEPARstart{G}{raph} anomaly detection aims to identify nodes that significantly deviate from the majority, and has drawn much attention from various domain~\cite{akoglu2015graph,kim2022graph}, such as fraud detection \cite{zhang2022efraudcom} and IoT network intrusion detection \cite{zhou2021hierarchical}.
Early works \cite{perozzi2016scalable,li2017radar,peng2018anomalous} on graph anomaly detection have been largely dependent on domain knowledge and statistical methods.
Recently, deep learning approaches have proved that they can effectively handle large-scale high-dimensional data and extract patterns from the data, thereby achieving satisfactory performance without the burden of handcrafting features~\cite{wang2019deep,langkvist2014review,kim2022graph}.
However, it could not effectively handle complex interactions and attribute information on attributed graph~\cite{su2022comprehensive}.
Graph neural networks (GNNs), more recently, have been adopted to efficiently and intuitively detect anomalies from graphs due to the highly expressive capability via the message passing mechanism in learning graph representations (e.g., \cite{ding2019deep,li2019specae}).
The message passing mechanism in GNNs treats all messages equally without considering their relative importance or relevance~\cite{kipf2016semi}. 
Frequently, particular messages or neighboring nodes may carry more valuable information~\cite{velivckovic2017graph,xiao2022graph}. 
Hence, efficiently capturing these important messages or nodes can be a key to an advancement for detection anomalies.

Existing methods are predominantly developed in an unsupervised manner for graph anomaly detection~\cite{perozzi2016scalable,li2017radar,peng2018anomalous,fan2020anomalydae,ding2019deep,li2019specae,liu2021anomaly,pei2021resgcn,luo2022comga} since labeling anomalies are labor-intensive and require specific domain knowledge. Among the existing methods, most approaches are based on graph autoencoder (GAE)~\cite{ding2019deep,li2019specae,peng2018anomalous,fan2020anomalydae,pei2021resgcn,luo2022comga}, which measures the abnormality of each node with its reconstruction errors.
Unfortunately, the detected anomalies may be found to be noises or data instances that we are not interested in and one of possible reasons can be the absence of the prior knowledge about anomalies we are looking for~\cite{ding2021few,ding2022meta,tian2023sad}.
Without the prior knowledge or anomaly-aware information, it may be less effective to obtain the results that we exactly want from unsupervised approaches~\cite{tian2023sad}.
The above issue can be solved by using a few or limited number of labeled anomalies as the prior information to catch anomaly-informed features~\cite{ma2021comprehensive,pang2021deep}.
In practice, it is often recommended to utilize as much labeled data as possible~\cite{aggarwal2017introduction}. Furthermore, in real-world scenarios, we might be able to collect a small set of labeled anomalies~\cite{ma2021comprehensive, ding2021few}.

By using few anomaly labeled data as prior knowledge, the issues from unsupervised approaches could be resolved~\cite{ma2021comprehensive}.
Even though how effectively using few labeled information is important, it still remains non-trivial mostly due to the following issues:
(1) \textit{Precise abnormality characterization}: since there is only limited amount of labeled information, it is difficult to accurately characterize the abnormal patterns~\cite{ding2021few}.
As the number of anomalies used are increased, the abnormal patterns are characterized more easily and precisely and previous semi-supervised methods~\cite{pang2019deep,kumagai2021semi} show performance enhancement by using more anomalies.
Nonetheless, due to the fact that labeling cost is high and it is hard to obtain enough labels in real world, a method for handling as few labeled anomalies as possible to learn high-level abstraction of abnormal patterns is necessary.
In addition, utilizing abundant unlabeled information is also helpful in learning the high-level abnormal features.
2) \textit{Capturing valuable representation}: how to efficiently capture valuable normality/abnormality representations with a small amount of labeled data is also one of the major challenges~\cite{pang2021deep}.
Various research studies~\cite{ruffdeep,kumagai2021semi,pang2019deep} attempt to distinguish the more important features and to leverage label information in semi-supervised manner.
The performance of these methods is prone to decrease as fewer labels are leveraged.
Hence, there is a need for a carefully designed model, which exhibits stable performance with fewer labels and captures representative characteristics of labeled and unlabeled information.

\begin{figure}[t]
    \centering
    \includegraphics[width=0.99\linewidth]{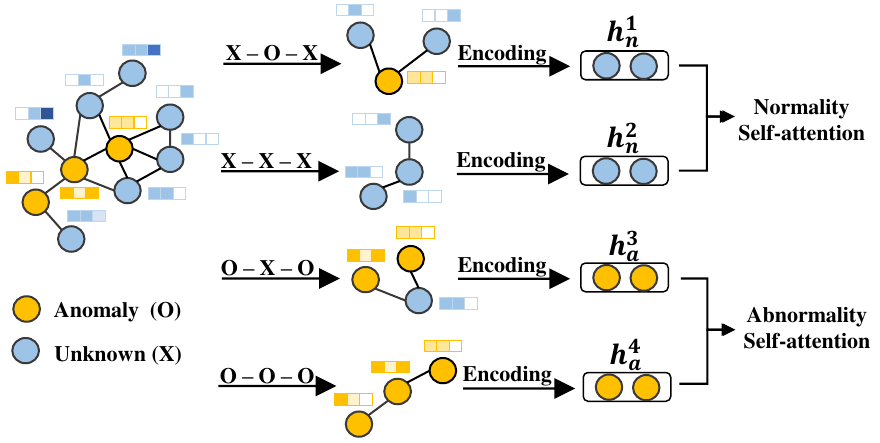}
    \caption{An example of how attention mechanism is achieved by using metapath-based context information.}
    \label{fig:self-attention}
\end{figure}

To alleviate aforementioned challenges, we first propose graph augmentation algorithm to 
embed actual linking patterns between normal and anomalous nodes in a graph by using both metapath schema and node label types (unknown or abnormal).
Particularly, based on label types, we sample anomaly subgraphs with designed metapath schema, which can enhance normality- and abnormality-specific information in the form of self-attention as shown in Figure \ref{fig:self-attention}.
Hence, the anomaly subgraphs contain higher-level context and relation information between each type of nodes.
Moreover, the procedures of sampling subgraphs effectively augment label augmentation and it helps reducing class imbalance problem as listed in Table \ref{tab:info_usage}.
To further effectively exploit the context information from the anomaly subgraphs, we introduce a well-designed framework Metapath-based Graph Anomaly Detection (MGAD).
Specifically, MGAD employs GNN-based graph autoencoder as its backbone network and dual encoders, which capture the complex interactions as well as metapath-based context information between labeled and unlabeled nodes from entire graph and anomaly subgraph.

\spara{Metapath-based context information}.
As shown in Figure \ref{fig:self-attention}, there are multiple metapaths, which represent different relations between unknown (blue) and/or abnormal (orange) nodes, such as Unknown-Anomaly-Unknown (X-O-X), Unknown-Unknown-Unknown (X-X-X), Anomaly-Unknown-Anomaly (O-X-O), and Anomaly-Anomaly-Anomaly (O-O-O).
Hence, O-X-O and O-O-O represent two different relations between anomalous nodes.
X-X-X and O-O-O express the patterns of unknown and anomalous nodes respectively.
X-O-X expresses the patterns of anomalies in unknown nodes, and O-X-O represents the patterns of unknown nodes in anomalies.
By incorporating multiple metapaths, features between unknown nodes and anomalous nodes become more distinguishable and we can leverage specific context information to detect anomalies.
In addition, attention mechanism allows a model to dynamically focus on relevant metapaths and adaptively weigh the importance of different metapaths based on its relevance, enabling more context-aware anomaly detection as described in Figure \ref{fig:self-attention}.
The metapaths and attention mechanism mechanism can lead to more accurate and effective detection performance compared to the previous GNN-based graph anomaly detection approaches.

In summary, our main contributions are listed below:
\begin{itemize}[leftmargin=*]
    \item To the best of our knowledge, the proposed method is the first attempt to adopt metapath-based context information and attention mechanism for graph anomaly detection. The combination of metapath-based context information and attention mechanism enables normal and abnormal representations to be more distinguishable.
    Also, it is first attempt to adopt metapath for homogeneous graph.
    \item We effectively augment label data by iterative sampling strategy of metapaths and this strategy alleviates class imbalance problem in graph anomaly detection.
    \item We conduct extensive experiments on seven real-world datasets to show the performance of our framework. The results demonstrate the superiority of our method by comparing with the state-of-the-art approaches.
\end{itemize}

\begin{table*}[t]
\centering
\caption{The usage of information of previous GNN-based methods for graph anomaly detection.}
\label{tab:info_usage}
\begin{tabular}{@{}llcccccc@{}}
\toprule
 &
  Methods &
  \begin{tabular}[c]{@{}c@{}}Local\\ Information\end{tabular} &
  \begin{tabular}[c]{@{}c@{}}Global\\ Information\end{tabular} &
  \begin{tabular}[c]{@{}c@{}}Context\\ Information\end{tabular} &
  \begin{tabular}[c]{@{}c@{}}Global\\ Anomaly\end{tabular} &
  \begin{tabular}[c]{@{}c@{}}Local\\ Anomaly\end{tabular} &
  \begin{tabular}[c]{@{}c@{}}Class\\ Imbalance\end{tabular} \\ \midrule
\multirow{14}{*}{Unsupervised}   & AMEN~\cite{perozzi2016scalable}        & $\surd$ & $\surd$ & $\times$ & $\surd$ & $\times$ & $\times$ \\
                                 & Radar~\cite{li2017radar}       & $\surd$ & $\surd$ & $\surd$ & $\surd$ & $\surd$ & $\times$ \\
                                 & ANOMALOUS~\cite{peng2018anomalous}   & $\surd$ & $\surd$ & $\times$ & $\surd$ & $\surd$ & $\times$ \\
                                 & DOMINANT~\cite{ding2019deep} &$\times$&$\surd$&$\times$&$\surd$&$\times$&$\times$ \\
                                 & SpecAE~\cite{li2019specae}   &$\times$&$\surd$&$\times$&$\surd$&$\surd$&$\times$ \\ 
                                 & HCM~\cite{huang2021hop}      &$\surd$&$\surd$&$\surd$&$\times$&$\times$&$\times$ \\ 
                                 & AnomalyDAE~\cite{fan2020anomalydae}  & $\times$ & $\surd$ & $\times$ & $\surd$ & $\times$ & $\times$ \\
                                 & ResGCN~\cite{pei2021resgcn}      & $\times$ & $\surd$ & $\times$ & $\surd$ & $\times$ & $\times$ \\
                                 & CoLA~\cite{liu2021anomaly}        & $\surd$ & $\times$ & $\surd$ & $\times$ & $\surd$ & $\surd$ \\
                                 & ConGNN~\cite{li2024controlled}      & $\times$ & $\surd$ & $\times$ & $\surd$ & $\times$ & $\surd$ \\
                                 & LHML~\cite{guo2022learning}        & $\surd$ & $\times$ & $\times$ & $\times$ & $\surd$ & $\times$ \\
                                 & ComGA~\cite{luo2022comga}    &$\surd$&$\surd$&$\times$&$\surd$&$\surd$&$\times$ \\
                                 & BWGNN~\cite{tang2022rethinking}       & $\times$ & $\surd$ & $\times$ & $\surd$ & $\times$ & $\times$ \\
                                 & GHRN~\cite{gao2023addressing}        & $\surd$ & $\surd$ & $\times$ & $\surd$ & $\times$ & $\surd$ \\ \midrule
\multirow{4}{*}{Semi-supervised} & DeepSAD~\cite{ruffdeep}     & $\times$ & $\surd$ & $\times$ & $\surd$ & $\times$ & $\times$ \\
                                 & GDN~\cite{ding2021few}         & $\times$ & $\surd$ & $\times$ & $\surd$ & $\times$ & $\times$ \\
                                 & Semi-GNN~\cite{kumagai2021semi}    & $\times$ & $\surd$ & $\times$ & $\surd$ & $\times$ & $\surd$ \\
                                 & MGAD (ours) & $\surd$ & $\surd$ & $\surd$ & $\surd$ & $\surd$ & $\surd$ \\ \bottomrule
\end{tabular}
\end{table*}

\section{RELATED WORK} \label{sec:related_work}
In this section, we present methods for graph anomaly detection, ranging from traditional methods to state-of-the-art GNN-based approaches, in both unsupervised and semi-supervised settings. 
Additionally, the summary of state-of-the-art methods is listed in Table \ref{tab:info_usage}.

\spara{Traditional Methods}.
The research area of graph anomaly detection has been attracting significant attention in recent times. 
The earlier methods \cite{breunig2000lof,xu2007scan} consider only the structure or attribute to detect anomalous nodes in a graph.
Afterward, CODA \cite{gao2010community} presents a probabilistic model to detect anomalous nodes on communities.
AMEN \cite{perozzi2016scalable} builds the ego-graphs with each node in order to detect anomalous communities on attributed graphs.
Additionally, other traditional approaches \cite{muller2013ranking,perozzi2014focused,sanchez2013statistical,sanchez2014local} attempt to detect anomalies by using attribute information.
Also, some research \cite{li2017radar,peng2018anomalous} employs residual analysis techniques and ANOMALOUS \cite{peng2018anomalous} combines the residual analysis and CUR decomposition in order to detect anomalous nodes. 

\spara{Unsupervised GNN-based Methods}.
With the advent of GNNs \cite{hamilton2017inductive,kipf2016semi,velivckovic2017graph}, various GNN-based approaches for graph anomaly detection have been presented with skyrocketing growth of GNN techniques.
DOMINANT \cite{ding2019deep} firstly adopts GCN-based GAE, which encodes structure and attribute information and then decodes the structure and attribute features respectively for anomalous node detection.
CoLA \cite{liu2021anomaly} proposes  contrastive self-supervised learning for anomalous node detection by sampling local subgraphs and comparing the subgraphs with target nodes.
However, these above methods may not effectively work on real-world networks since the patterns of anomalies are much more irregular.
SpecAE \cite{li2019specae} attempts to detect global and community anomalies by using Laplacian sharpening technique to alleviate over-smoothing problem of convolution mechanism.
HCM \cite{huang2021hop} considers both global and local contextual anomalies on the basis of hop count-based subgraphs as self-supervised manner.
ComGA \cite{luo2022comga} captures global, local, and structure anomalies by utilizing structure information of community structure, attribute and structure features of the whole graph.

\spara{Semi-supervised GNN-based Methods}.
Several methods using graphs in semi-supervised way have been presented and these methods propagate labelled information to unlabelled nodes ~\cite{sen2008collective,zhou2003learning,zhu2003semi}.
Although GNN-based node classification methods in a semi-supervised way have achieved huge success~\cite{hamilton2017inductive,kipf2016semi,wu2019simplifying,yang2016revisiting}, these methods do not consider the class imbalance issue in the graph anomaly detection task.
There are only a few GNN-based approaches detecting anomalies in semi-supervised manner.
Kumagai et al. present a simple GCN-based embedding method in semi-supervised manner for graph anomaly detection~\cite{kumagai2021semi}.
Meta-GDN~\cite{ding2021few} introduce Graph Deviation Networks (GDN), which leverages a few labels to deviate abnormal nodes from normal nodes, with cross-network meta-learning algorithm.
In this paper, we propose GNN-based anomaly detection method in semi-supervised manner with metapath-based context information.

\section{PROBLEM STATEMENT} \label{sec:problem_statement}
In this section, we formally define key concepts and present the problem statement.

\begin{defn} \textbf{Attributed Graph}. \normalfont An attributed graph can be denoted as $\mathcal{G} = (V, E, X)$ where $V$ is a set of nodes $V = \{v_1, v_2, \cdots, v_n\}$ with $|V| = n$, $E$ is a set of edges with $|E| = m$, and $X \in \mathbb{R}^{n \times d}$ denotes node attributes.
The $i$-th row of vector $x_i \in \mathbb{R}^{d} (i = 1, \dots, n)$ is the attribute information of the node $v_i$.
Additionally, adjacency matrix $A$, where $A_{ij} = 1$ denotes connectivity between node $v_i$ and $v_j$, if $A_{ij} = 0$, then there is no connectivity each other.
\end{defn}

\begin{defn} \textbf{Metapath}. \normalfont A metapath $P$ is a path in the form of $V_1 \overset{R_1}{\rightarrow} V_2 \overset{R_2}{\rightarrow} \cdots \overset{R_l}{\rightarrow} V_{l+1}$, which defines a composite relation $R = R_1 \circ R_2 \circ \cdots \circ R_l$ between types of nodes $V_1$ and $V_{l+1}$, where $\circ$ denotes the composition operator on relations.
\end{defn}

\begin{defn} \textbf{Metapath-based Anomaly Subgraph.} \label{def:anomaly_subgraph} 
\normalfont A metapath based anomaly subgraph is a set of sampled metapath $\mathbf{T}_\mathbf{S} = (\mathbf{A_S}, \mathbf{R_S})$ on given metapath schema $\mathbf{T}_\mathcal{G} = (\mathbf{A}, \mathbf{R})$, where $\mathbf{A_S} \subseteq \mathbf{A}$ and $\mathbf{R_S} \subseteq \mathbf{R}$. Types of an anomaly subgraph $\mathbf{T_S}$ are either anomaly type O or unknown type X. The anomaly subgraph is recursively sampled with random walk sampling strategy.
\end{defn}

As graph anomaly detection is commonly formulated as a ranking problem~\cite{akoglu2015graph}, we formally define the metapath-based graph anomaly detection problem as follows:

\begin{prob} \textbf{Metapath-based Anomaly Detection.}
    \normalfont
    Given an attributed graph $\mathcal{G} = (V, E, X)$ with a set of nodes $V = \{v_1, v_2, \cdots, v_n\}$, the task is to score each node with a scoring function $k_i = f(v_i)$ by preserving the metapath-based context information.
    The anomaly score $k_i$ is the degree of abnormality of node $v_i$.
    Based on the score, anomalous nodes can be detected in accordance with the ranking.
\end{prob}

\section{THE PROPOSED APPROACH} \label{sec:proposed approach}
\begin{figure}[t]
    \centering
    \includegraphics[width=0.999\linewidth]{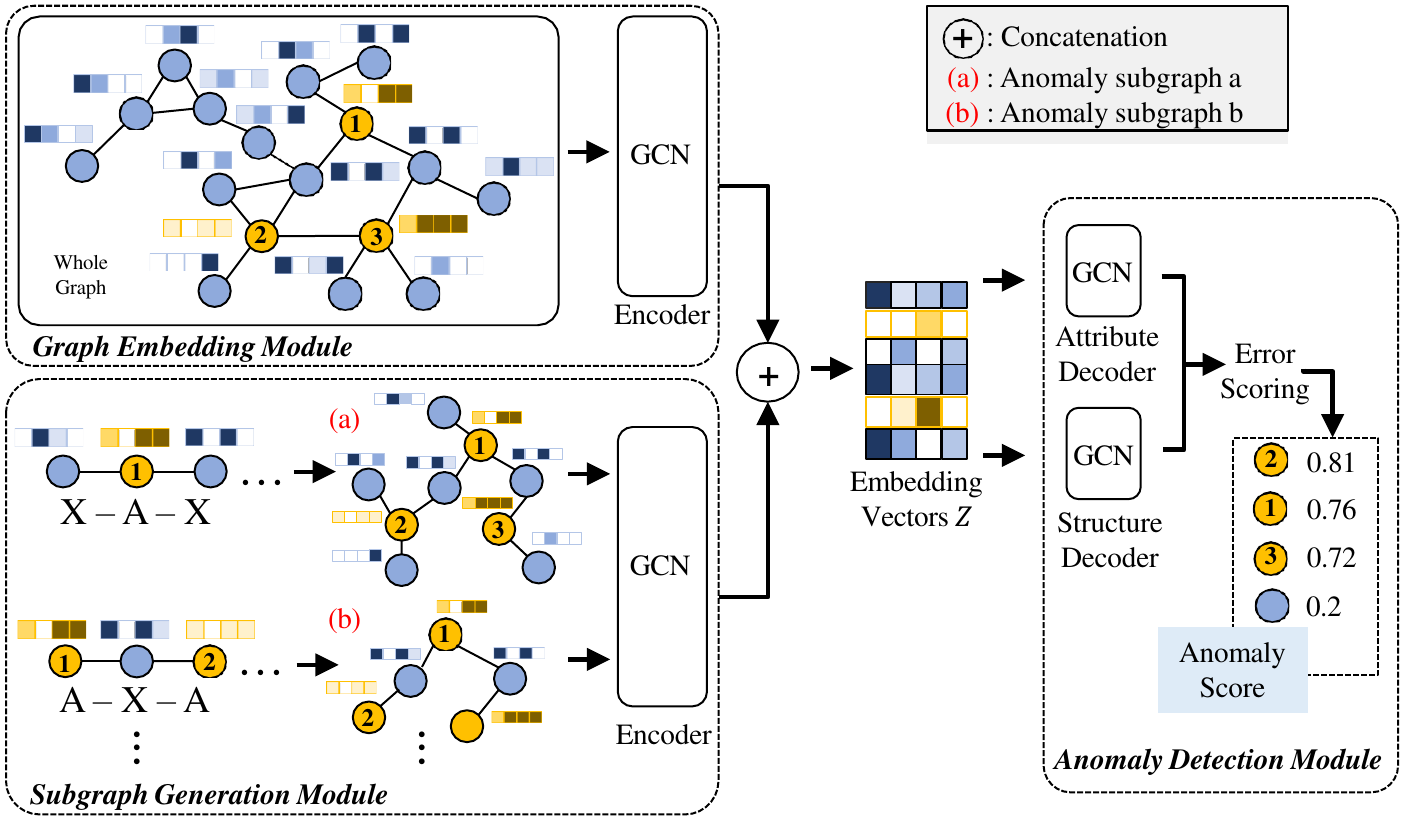}
    \caption{Framework of MGAD.}
    \label{fig:MSAD}
\end{figure}

Here, we present three modules of our framework: (1) Subgraph Generation Module, (2) Graph Embedding Module, and (3) Anomaly Detection Module, as in Figure \ref{fig:MSAD}.

\subsection{Subgraph Generation Module} 
In this module, the anomaly subgraph is generated on the basis of node types of O and X with metapath schema. Afterward, the subgraph is encoded by multiple GCN layers to extract important information, such as O-X relations, structure, and attribute features in the subgraph. 

Specifically, some of labelled anomaly nodes are set to type O and the others of the remaining nodes are set to type X since it is uncertain which node is either anomalous or not.
On the basis of metapath schema, which contains at least one O type, each node is randomly sampled by random walk sampling strategy.
The sampled nodes will be an anomaly subgraph, which represents normal-abnormal patterns as well as structure, attribute information in the subgraph as defined in Definition \ref{def:anomaly_subgraph}.
For example, given a metapath schema, X-O-X, every matching pattern is sampled from whole graph. If there are many links between nodes, then a random walker will randomly pick one node among linked nodes with probability $\mathbf{P}$.
The randomwalk-based sampling can be described as follows:
\begin{equation} \label{eq:randomwalk}
    \mathbf{P_{A_{l},A_{l+1}}} = \mathbf{D^{-1}_{A_{l},A_{l+1}}} \mathbf{W_{A_{l},A_{l+1}}}
\end{equation}
where $\mathbf{W_{A_{l},A_{l+1}}}$ is the adjacency matrix between nodes in type $\mathbf{A_l}$ and nodes in type $\mathbf{A_{l+1}}$, and $\mathbf{D_{A_{l},A_{l+1}}}$ is the degree matrix with $\mathbf{D_{A_{l},A_{l+1}}} = \sum \mathbf{W_{A_{l},A_{l+1}}}(v_i,v_j)$. 
A random walker samples each type of node on the basis of the probability $\mathbf{P}$ following the number of sampling round $n$.
For example, if a random walker starts sampling from a node $v_i$ in type $\mathbf{A_l}$, the next node will be $v_j$ in type $\mathbf{A_{l+1}}$ with the probability $\mathbf{P_{A_{l},A_{l+1}}}(v_i,v_j)$.
The entire procedures of subgraph generation are described in Algorithm \ref{alg:anomaly_subgraph}.
Specifically, we iteratively extract sub-structures, which match with designed metapath schema.
In line 5, we select one of the neighbors from target node uniformly at random.

Note that the proposed anomaly subgraph guarantees that there is at least one anomaly in each subgraph and it sufficiently represents anomalous patterns between normal and abnormal.
These information in subgraph will be efficiently encoded by multi-stacked GCN layers and will enhance anomaly detection performance by making normal and abnormal more distinguishable. Each GCN layer can be shown as follows:
\begin{equation} \label{eq:gcn} 
    \mathbf{H}^{(l)}_{i} = \phi(\tilde{\mathbf{D}}^{-\frac{1}{2}}_{i}\tilde{\mathbf{A}}_{i}\tilde{\mathbf{D}}^{-\frac{1}{2}}_{i}\mathbf{H}^{(l-1)}_{i}\mathbf{W}^{(l-1)})
\end{equation}
where $\tilde{\mathbf{A}}_i = \mathbf{A}_i + \mathbf{I}$ is the subgraph adjacency with self-loop, $\tilde{\mathbf{D}}_i$ is the degree matrix of local subgraph, $\mathbf{W}^{l-1}$ is the weight matrix of the $(l-1)$-th layer, and $\phi$ is the activation function, such as ReLU and Sigmoid.

\subsection{Graph Embedding Module}
In this module, GCN-based encoder, which is the same as the encoder in subgraph Generation Module, embeds structure and attribute information of the attributed graph.
The multiple GCN layers in the encoder extract global attribute and structure feature.

Specifically, the GCN layers capture non-linearity of network and complex relations on attributed networks as described in Equation \ref{eq:gcn}.
The embedding representation $\mathbf{Z}^{(l)}$ is concatenated with representation $\mathbf{H}^{(l)}$ of embedded anomaly subgraph and it composes embedding vector $\mathbf{Z}$, which consists of X-O linking patterns and information of node attribute and structure in anomaly subgraph and whole graph as described in Equation \ref{eq:embedding vector}.
\begin{equation} \label{eq:embedding vector}
    \mathbf{Z} = \mathbf{Z}^{(l)} \oplus \mathbf{H}^{(l)}
\end{equation}
Consequently, the embedding vector $\mathbf{Z}$ represents global, local, and X-O connecting features as well as attribute and structure characteristics.

\subsection{Anomaly Detection Module}
In this module, two GCN-based decoders reconstruct structures and attributes embedding vector $\mathbf{Z}$ respectively. Subsequently, anomalous nodes are detected on the basis of anomaly scores.

\SetKwInOut{Input}{input}
\RestyleAlgo{ruled}
\SetKwData{false}{false}
\SetKwData{true}{true}
\SetKwFunction{randomWalk}{rw}
\SetKwFunction{append}{append}
\SetKwFunction{type}{type}
\begin{algorithm}[t]
\caption{Anomaly subgraph generation} \label{alg:anomaly_subgraph}
\Input{
    The attributed graph $\mathcal{G} = (V, E, X)$, \\
    \# walks per node $n$, \\
    Walk length $l$, \\
    Metapath schema $P$
}
\KwOut{Anomaly subgraph $C$}
$C \gets \emptyset$\;
\For{$v \in V$}{
    $i \leftarrow 0$\;
    \While{$i \neq n$}{
        $H \leftarrow$ \randomWalk{$v$, $\mathcal{G}$, $l$}\;
        \If{\type{$H$} $\in P$}{
            \append{$C$, $H$}\;
        }
        $i \leftarrow i + 1$\;
    }
}
\Return{$C$}
\end{algorithm}

\spara{Structure Decoder.}
In this section, we explain how the structure decoder reconstructs the original graph structure with latent representations $\mathbf{Z}$.
The reconstruction error of structure is described as $\mathbf{R}_{struc} = \mathbf{A} - \hat{\mathbf{A}}$, where $\mathbf{\hat{A}}$ denotes estimated adjacency matrix.
Additionally, if a node shows low probability of being anomalous, the structure decoder well reconstructs topological information of the graph.
On the other hand, if the decoder properly reconstructs the topological information, then its topology does not follow majority structures in a graph.
Thus, $\mathbf{R}_{struc(i,:)}$ expresses that $i^{th}$ node has a higher anomaly probability on the perspective of network structure.
We use structure decoder with the latent representation $\mathbf{Z}$ as input to reconstruct the graph structure:
\begin{equation} \label{eq:structure_decoder_2}
    \hat{\mathbf{A}} = sigmoid(\mathbf{ZZ^T})
\end{equation}
Specifically, given the latent representations as input, the decoder predicts whether there is a link between each pair of two nodes as follows:
\begin{equation} \label{eq:structure_decoder_1}
    p(\hat{\mathbf{A}}_{i,j} = 1|\mathbf{z}_i,\mathbf{z}_j) = sigmoid(\mathbf{z}_i \mathbf{z}^{\mathbf{T}}_{j})
\end{equation}

\spara{Attribute Decoder.}
In this section, we describe how this attribute decoder reconstructs the original node attributes of the graph with $\mathbf{Z}$.
The reconstruction error of attribute is expressed as $\mathbf{R}_{attr} = \mathbf{X - \hat{X}}$, where $\mathbf{\hat{X}}$ denotes predicted attributes.
The attribute decoder employs GCN layers to predict the original attributes as follows:
\begin{equation} \label{eq:attribute_decoder}
    \hat{\mathbf{X}} = \mathbf{\phi}(\mathbf{Z,A,W})
\end{equation}

\spara{Anomaly Detection.}
In order to learn the structure and attribute reconstruction errors, the objective function can be defined as:
\begin{equation} \label{eq:objective_function}
    \mathbf{L} = (1-\alpha)||\mathbf{A - \hat{A}||^{2}_{F}} + \alpha\mathbf{||X - \hat{X}||^{2}_{F}}
\end{equation}
where $\alpha$ is a balance parameter, which controls importance between structure and attribute reconstruction errors.
The joint reconstruction errors are used to measure anomalous nodes by minimizing the objective function above.
The anomaly score of each node $\mathbf{v}_i$ is described as follows:
\begin{equation}
    score(\mathbf{v}_i) = (1 - \alpha)||\mathbf{a} - \mathbf{\hat{a}}_i||_{2} + \alpha||\mathbf{x}_i - \mathbf{\hat{x}}_i||_2
\end{equation}
Note that we simply consider the unlabeled nodes $V$ as normal nodes~\cite{pang2019deep,ding2021few}. 
Outstandingly, MGAD shows novel and robust performance on this straightforward strategy.

\subsection{Time Complexity Analysis}

The time complexity of generating Metapath-based anomaly subgraph is $O(|V| \cdot n \cdot l)$. 
The outer loop runs for each node $v$ in the graph $\mathcal{G}$, resulting in a time complexity of $O(|V|)$, where $|V|$ is the number of nodes in the graph.
The inner loop runs $n$ times ($1 \leq n \leq 5$) for each node $v$. Inside the inner loop:
\begin{itemize}[leftmargin=*]
    \item A random walk $H$ is generated using the function $rw(v, \mathcal{G}, l)$, where $l$ is the walk length ($3 \leq l \leq 5$).
    \item The type of the walk $H$ is checked against the metapath schema $P$, and if it matches, the walk is added to anomaly subgraph $C$. 
\end{itemize}
Generating a random walk using the $rw(v, \mathcal{G}, l)$ function involves traversing the graph with a specified length $l$. 
Assuming the time complexity of generating a random walk is $O(l)$ in the worst case, and the metapath type check is constant time (since the metapath schema is typically a small fixed-size set of metapaths), the overall time complexity of the inner loop is $O(n * l)$.
Therefore, the overall time complexity of the algorithm is $O(|V| \cdot n \cdot l)$.

\begin{table}[t]
\centering
\caption{The statistics of the datasets.}
\resizebox{0.85\columnwidth}{!}{%
\begin{tabular}{@{}c|l|cccc@{}}
\toprule
\multicolumn{1}{c|}{}              & Datasets    & $|V|$  & $|E|$   & $|X|$  & $|A|$\phantom{1} \\ \midrule
Ground-truth                      & Amazon      & 1,418  & 3,695   & 28     & 28\phantom{1}    \\ \midrule
\multirow{6}{*}{\begin{tabular}[c]{@{}c@{}}Injected \\ Anomaly\end{tabular}} & Cora        & 2,708  & 5,429   & 1,433  & 150\phantom{1}   \\
                                  & Citeseer    & 3,327  & 4,732   & 3,703  & 150\phantom{1}   \\
                                  & Pubmed      & 19,717 & 44,338  & 500    & 600\phantom{1}   \\
                                  & BlogCatalog & 5,196  & 171,742 & 8,189  & 300\phantom{1}   \\
                                  & Flickr      & 7,575  & 239,738 & 12,074 & 450\phantom{1}   \\
                                  & ACM         & 16,484 & 71,980  & 8,337  & 600\phantom{1}  \\ \bottomrule
\end{tabular}%
}
\label{tab:datasets}
\end{table}

\begin{table}[t]
\centering
\caption{AUC values (\%) on seven datasets. The best and second highest performances are boldfaced and underlined, respectively.}
\label{tab:AUC_results}
\resizebox{\columnwidth}{!}{%
\begin{tabular}{@{}lcccccc@{}}
\toprule
  Methods &
  Cora &
  Citeseer &
  Pubmed &
  BlogCatalog &
  Flickr &
  ACM \\ \midrule

  AMEN &
  47.00 &
  62.66 &
  61.54 &
  77.13 &
  65.73 &
  56.26 \\
 Radar      & 65.87 & 67.09 & 62.33          & 74.01          & 73.99          & 72.47 \\
 ANOMALOUS  & 57.70 & 63.07 & 73.16          & 72.37          & 74.34          & 70.38 \\
 DOMINANT   & 81.55 & 82.51 & 80.81          & 74.68          & 74.42          & 76.01 \\
 AnomalyDAE & 76.27 & 72.71 & 81.03          & 78.34          & 75.08          & 75.13 \\
 ResGCN     & 84.79 & 76.47 & 80.79          & 78.50          & 78.00          & 76.80 \\
 CoLA       & 87.79 & 89.68 & \textbf{95.12} & 78.54          & 75.13          & 82.37 \\
 ComGA      & 88.40 & \underline{91.67} & 92.20          & 81.40          & 79.90          & \underline{84.96} \\
 ConGNN     & \underline{89.05} & OOM   & 87.26          & OOM            & OOM            & OOM   \\
 LHML       & 84.40 & 80.20 & 83.50          & 72.60          & 70.60          & 82.10 \\
 BWGNN      & 48.63 & 43.11 & 70.60          & 42.57          & 78.91          & 64.71 \\
 GHRN       & 67.91 & 57.60 & 79.07          & 80.42          & \textbf{83.06} & 75.52 \\ \midrule

  DeepSAD &
  56.70 &
  54.60 &
  51.10 &
  54.80 &
  53.90 &
  53.00 \\
 GDN        & 83.17 & 85.48 & 84.72          & 70.90          & 71.03          & 78.12 \\
 Semi-GNN   & 67.65 & 81.74 & 66.30          & \textbf{84.42} & 73.39          & 84.90 \\
 MGAD (ours) &
  \textbf{92.29} &
  \textbf{94.69} &
  \underline{92.52} &
  \underline{81.48} &
  \underline{79.96} &
  \textbf{89.87} \\ \bottomrule
\end{tabular}%
}
\end{table}

\begin{figure*}[t]
\pgfplotsset{every axis title/.append style={at={(0.5,0.9)}}}
\begin{tikzpicture}
\begin{groupplot}[group style={group size=7 by 1}, 
    height=3.2cm,
    width=0.175\textwidth,
    axis y line*=left,
    axis x line*=left,
    ylabel near ticks,
    xlabel near ticks,
    xtick={1, 3, 5},
    xmin=0,
    xmax=7,
    xticklabels={3, 4, 5},
]
\nextgroupplot[title=Amazon, ylabel=AUC]
\addplot[
    scatter/classes={a={orange}, b={orange}, c={orange}},
    scatter,
    only marks,
    scatter src=explicit symbolic,
    ]%
    plot [error bars/.cd, y dir = both, y explicit]
    table [meta=class, x=x, y=avg, y error plus=max, y error minus=min] {
        x avg max min class
        1 63.94 0.3 0.51 a
        3 63.76 0.68 0.46 b
        5 63.95 1.05 0.68 c
    };

\nextgroupplot[title=ACM]
\addplot[
    scatter/classes={a={orange}, b={orange}, c={orange}},
    scatter,
    only marks,
    scatter src=explicit symbolic,
    ]%
    plot [error bars/.cd, y dir = both, y explicit]
    table [meta=class, x=x, y=avg, y error plus=max, y error minus=min] {
        x avg max min class
        1 89.73 0.13 0.07 a
        3 89.7 0.12 0.66 b
        5 89.71 0.15 0.12 c
    };

\nextgroupplot[title=Cora]
\addplot[
    scatter/classes={a={orange}, b={orange}, c={orange}},
    scatter,
    only marks,
    scatter src=explicit symbolic,
    ]%
    plot [error bars/.cd, y dir = both, y explicit]
    table [meta=class, x=x, y=avg, y error plus=max, y error minus=min] {
        x avg max min class
        1 91.44 0.67 1.29 a
        3 91.45 0.61 1.3 b
        5 91.52 0.76 1.07 c
    };

\nextgroupplot[title=Citeseer]
\addplot[
    scatter/classes={a={orange}, b={orange}, c={orange}},
    scatter,
    only marks,
    scatter src=explicit symbolic,
    ]%
    plot [error bars/.cd, y dir = both, y explicit]
    table [meta=class, x=x, y=avg, y error plus=max, y error minus=min] {
        x avg max min class
        1 93.74 0.94 0.72 a
        3 93.74 0.82 0.61 b
        5 93.65 0.66 0.7 c
    };
    
\nextgroupplot[title=Pubmed]
\addplot[
    scatter/classes={a={orange}, b={orange}, c={orange}},
    scatter,
    only marks,
    scatter src=explicit symbolic,
    ]%
    plot [error bars/.cd, y dir = both, y explicit]
    table [meta=class, x=x, y=avg, y error plus=max, y error minus=min] {
        x avg max min class
        1 92.15 0.36 0.44 a
        3 92.03 0.42 0.38 b
        5 92.03 0.39 0.5 c
    };
    
\nextgroupplot[title=BlogCatalog]
\addplot[
    scatter/classes={a={orange}, b={orange}, c={orange}},
    scatter,
    only marks,
    scatter src=explicit symbolic,
    ]%
    plot [error bars/.cd, y dir = both, y explicit]
    table [meta=class, x=x, y=avg, y error plus=max, y error minus=min] {
        x avg max min class
        1 81.43 0.01 0.0 a
        3 81.45 0.02 0.01 b
        5 81.45 0.01 0.0 c
    };
    
\nextgroupplot[title=Flickr]
\addplot[
    scatter/classes={a={orange}, b={orange}, c={orange}},
    scatter,
    only marks,
    scatter src=explicit symbolic,
    ]%
    plot [error bars/.cd, y dir = both, y explicit]
    table [meta=class, x=x, y=avg, y error plus=max, y error minus=min] {
        x avg max min class
        1 79.82 0.12 0.02 a
        3 79.81 0.02 0.0 b
        5 79.81 0.01 0.01 c
    };
    
\end{groupplot}
\end{tikzpicture}
\centering
\caption{AUC values on different length of metapath $l$.}
\label{fig:results_length}
\end{figure*}
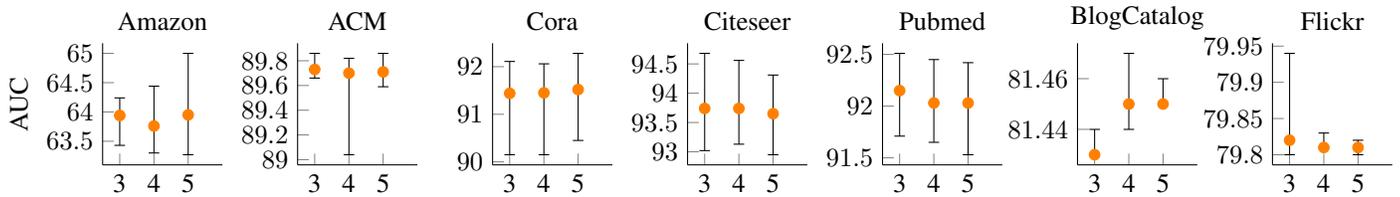
\begin{figure*}[t]
\pgfplotsset{every axis title/.append style={at={(0.5,0.9)}}}
\begin{tikzpicture}
\begin{groupplot}[group style={group size=7 by 1}, 
    height=3.2cm, 
    width=0.175\textwidth,
    axis y line*=left,
    axis x line*=left,
    ylabel near ticks,
    xlabel near ticks,
    xtick={1, 3, 5},
    xmin=0,
    xmax=7,
    xticklabels={1, 3, 5},
]
\nextgroupplot[title=Amazon, ylabel=AUC]
\addplot[
    scatter/classes={a={cyan}, b={cyan}, c={cyan}},
    scatter,
    only marks,
    scatter src=explicit symbolic,
    ]%
    plot [error bars/.cd, y dir = both, y explicit]
    table [meta=class, x=x, y=avg, y error plus=max, y error minus=min] {
        x avg max min class
        1 63.89 0.6 0.51 a
        3 63.81 0.65 0.54 b
        5 63.98 1.03 0.68 c
    };

\nextgroupplot[title=ACM]
\addplot[
    scatter/classes={a={cyan}, b={cyan}, c={cyan}},
    scatter,
    only marks,
    scatter src=explicit symbolic,
    ]%
    plot [error bars/.cd, y dir = both, y explicit]
    table [meta=class, x=x, y=avg, y error plus=max, y error minus=min] {
        x avg max min class
        1 89.72 0.09 0.07 a
        3 89.7 0.12 0.66 b
        5 89.72 0.15 0.13 c
    };

\nextgroupplot[title=Cora]
\addplot[
    scatter/classes={a={cyan}, b={cyan}, c={cyan}},
    scatter,
    only marks,
    scatter src=explicit symbolic,
    ]%
    plot [error bars/.cd, y dir = both, y explicit]
    table [meta=class, x=x, y=avg, y error plus=max, y error minus=min] {
        x avg max min class
        1 91.38 0.72 1.23 a
        3 91.55 0.74 1.4 b
        5 91.54 0.58 0.96 c
    };

\nextgroupplot[title=Citeseer]
\addplot[
    scatter/classes={a={cyan}, b={cyan}, c={cyan}},
    scatter,
    only marks,
    scatter src=explicit symbolic,
    ]%
    plot [error bars/.cd, y dir = both, y explicit]
    table [meta=class, x=x, y=avg, y error plus=max, y error minus=min] {
        x avg max min class
        1 93.62 0.67 0.61 a
        3 93.62 0.69 0.52 b
        5 93.83 0.86 0.88 c
    };
    
\nextgroupplot[title=Pubmed]
\addplot[
    scatter/classes={a={cyan}, b={cyan}, c={cyan}},
    scatter,
    only marks,
    scatter src=explicit symbolic,
    ]%
    plot [error bars/.cd, y dir = both, y explicit]
    table [meta=class, x=x, y=avg, y error plus=max, y error minus=min] {
        x avg max min class
        1 92.03 0.42 0.51 a
        3 92.05 0.38 0.4 b
        5 92.05 0.46 0.35 c
    };
    
\nextgroupplot[title=BlogCatalog]
\addplot[
    scatter/classes={a={cyan}, b={cyan}, c={cyan}},
    scatter,
    only marks,
    scatter src=explicit symbolic,
    ]%
    plot [error bars/.cd, y dir = both, y explicit]
    table [meta=class, x=x, y=avg, y error plus=max, y error minus=min] {
        x avg max min class
        1 81.44 0.02 0.0 a
        3 81.44 0.0 0.0 b
        5 81.45 0.02 0.02 c
    };
    
\nextgroupplot[title=Flickr]
\addplot[
    scatter/classes={a={cyan}, b={cyan}, c={cyan}},
    scatter,
    only marks,
    scatter src=explicit symbolic,
    ]%
    plot [error bars/.cd, y dir = both, y explicit]
    table [meta=class, x=x, y=avg, y error plus=max, y error minus=min] {
        x avg max min class
        1 79.82 0.13 0.02 a
        3 79.82 0.1 0.02 b
        5 79.82 0.02 0.05 c
    };
    
\end{groupplot}
\end{tikzpicture}
\centering
\caption{AUC values by sampling round $n$.}
\label{fig:results_sampling_rounds}
\end{figure*}
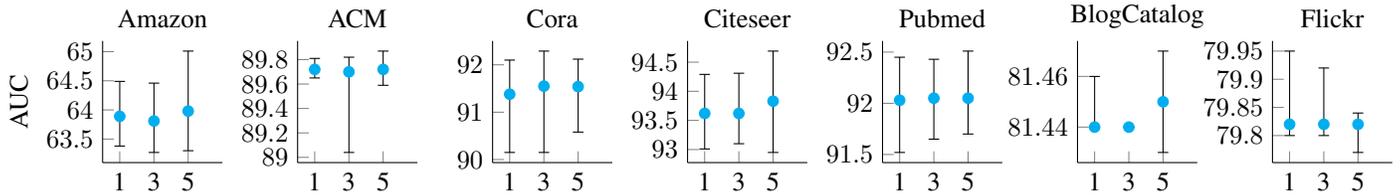

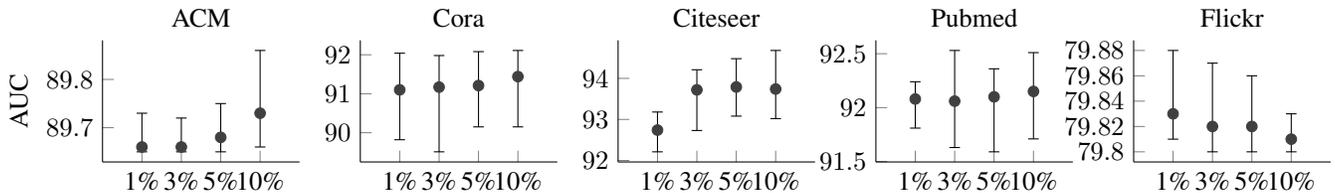
\begin{figure*}[t]
\pgfplotsset{every axis title/.append style={at={(0.5,0.9)}}}
\begin{tikzpicture}
\begin{groupplot}[group style={group size=5 by 1, horizontal sep=23pt}, 
    height=3.2cm,
    width=4.2cm,
    axis y line*=left,
    axis x line*=left,
    ylabel near ticks,
    xlabel near ticks,
    xtick={1, 3, 5, 7},
    xmin=-1,
    xmax=9,
    xticklabels={1\%, 3\%, 5\%, 10\%},
]

\nextgroupplot[title=ACM, ylabel=AUC]
\addplot[
    scatter/classes={a={darkgray}, b={darkgray}, c={darkgray}, d={darkgray}},
    scatter,
    only marks,
    scatter src=explicit symbolic,
    ]%
    plot [error bars/.cd, y dir = both, y explicit]
    table [meta=class, x=x, y=avg, y error plus=max, y error minus=min] {
        x avg max min class
        1 89.66 0.07 0.01 a
        3 89.66 0.06 0.01 b
        5 89.68 0.07 0.03 c
        7 89.73 0.13 0.07 d 
    };

\nextgroupplot[title=Cora]
\addplot[
    scatter/classes={a={darkgray}, b={darkgray}, c={darkgray}, d={darkgray}},
    scatter,
    only marks,
    scatter src=explicit symbolic,
    ]%
    plot [error bars/.cd, y dir = both, y explicit]
    table [meta=class, x=x, y=avg, y error plus=max, y error minus=min] {
        x avg max min class
        1 91.1 0.94 1.28 a
        3 91.17 0.81 1.66 b
        5 91.21 0.87 1.06 c
        7 91.44 0.67 1.29 d
    };

\nextgroupplot[title=Citeseer]
\addplot[
    scatter/classes={a={darkgray}, b={darkgray}, c={darkgray}, d={darkgray}},
    scatter,
    only marks,
    scatter src=explicit symbolic,
    ]%
    plot [error bars/.cd, y dir = both, y explicit]
    table [meta=class, x=x, y=avg, y error plus=max, y error minus=min] {
        x avg max min class
        1 92.74 0.44 0.53 a
        3 93.72 0.49 0.99 b
        5 93.79 0.69 0.71 c
        7 93.74 0.94 0.72 d
    };
    
\nextgroupplot[title=Pubmed] 
\addplot[
    scatter/classes={a={darkgray}, b={darkgray}, c={darkgray}, d={darkgray}},
    scatter,
    only marks,
    scatter src=explicit symbolic,
    ]%
    plot [error bars/.cd, y dir = both, y explicit]
    table [meta=class, x=x, y=avg, y error plus=max, y error minus=min] {
        x avg max min class
        1 92.08 0.16 0.27 a
        3 92.06 0.47 0.43 b
        5 92.1 0.26 0.51 c
        7 92.15 0.36 0.44 d
    };
    
\nextgroupplot[title=Flickr]
\addplot[
    scatter/classes={a={darkgray}, b={darkgray}, c={darkgray}, d={darkgray}},
    scatter,
    only marks,
    scatter src=explicit symbolic,
    ]%
    plot [error bars/.cd, y dir = both, y explicit]
    table [meta=class, x=x, y=avg, y error plus=max, y error minus=min] {
        x avg max min class
        1 79.83 0.05 0.02 a
        3 79.82 0.05 0.02 b
        5 79.82 0.04 0.02 c
        7 79.81 0.02 0.01 d
    };
    
\end{groupplot}
\end{tikzpicture}
\centering
\caption{Results following the number of anomalies (ratio of anomaly).}
\label{fig:results_number_of_anomalies}
\end{figure*}

\section{EXPERIMENTS} \label{sec:experiments}
In this section, we conduct an extensive experiment on seven real-world networks to demonstrate the superiority of MGAD.

\spara{Datasets}.
Among the seven datasets, only Amazon dataset provides ground-truth label and the other six datasets provide injected anomalies.
Previous methods for graph anomaly detection mostly utilized the injected anomaly networks for their experiments \cite{ding2021inductive,ding2019deep,fan2020anomalydae,li2019specae,liu2021anomaly} since the ground-truth information is rare in real-world datasets \cite{huang2021hop,pei2021resgcn,peng2018anomalous}.
Since there is no ground-truth in other datasets, we employ two anomaly injection techniques used in previous research \cite{liu2021anomaly,song2007conditional} to generate a comprehensive set of anomalies for each dataset by perturbing topological structure and node attributes, respectively.
Specifically, when injecting attribute anomalies, we randomly choose $n$ nodes and then change their features with randomly selected node's features.
Afterward, the equal number of structural anomalies are injected by selecting $n$ nodes and forming the chosen nodes fully connected.
To demonstrate the performance of MGAD, we conduct experiments on one ground-truth network and six injected anomaly networks~\cite{ding2019deep,liu2021anomaly,ding2021few}.
The statistics of these networks are listed in Table \ref{tab:datasets}, and more details are given as follows:
\begin{itemize}[leftmargin=*]
    \item Ground-truth anomaly graphs \cite{huang2021hop,pei2021resgcn,peng2018anomalous}: 
    Amazon is a co-purchase network, and an attributed network with ground-truth. The nodes labelled as amazonfail are represented as anomalous nodes.
    \item Injected anomaly graphs \cite{ding2021inductive,ding2019deep,fan2020anomalydae,li2019specae,liu2021anomaly,ding2021few}: 
    ACM, Cora, Citeseer, Pubmed, BlogCatalog, and Flickr are the six attributed networks with the injected anomalies.
    The first four datasets (ACM, Cora, Citeseer, and Pubmed) are widely used citation networks.
    Each node represents paper, and each edge expresses citation relation with other papers, respectively.
    The title and abstract information of papers are used as attributes of the nodes.
    BlogCatalog and Flickr are the two attributed networks and common social networks.
    Each node denotes the website user, and each edge represents the relations between users, respectively.
    User-related tag information, such as posting blogs or sharing photos is employed as node attributes.
\end{itemize}

\spara{Baselines}.
We introduce eight famous unsupervised and three semi-supervised methods for graph anomaly detection to compare with our proposed framework MGAD.
\begin{itemize}[leftmargin=*]
    \item AMEN~\cite{perozzi2016scalable} extracts both attribute and structure features to detect anomalies in ego-network by comparing correlations of nodes.
    \item Radar~\cite{li2017radar} analyzes residuals and coherence of network information for graph anomaly detection.
    \item ANOMALOUS~\cite{peng2018anomalous} uses CUR decomposition and residual to detect node anomalies.
    \item DOMINANT~\cite{ding2019deep} is based on GAE to extract node representation for graph anomaly detection.
    \item AnomalyDAE~\cite{fan2020anomalydae} extracts structure and attribute features by using two autoencoders.
    \item ResGCN~\cite{pei2021resgcn} adopts attention mechanism in residuals to detect anomalous node with GCN.
    \item CoLA~\cite{liu2021anomaly} is a GNN-based framework using contrastive learning to learn relations between target nodes and subgraphs.
    \item HCM~\cite{huang2021hop} is a novel self-supervised learning method with hop-based sub-structures of graphs.
    \item DeepSAD~\cite{ruffdeep} is a deep learning-based method for semi-supervised anomaly detection.
    \item GDN~\cite{ding2021few} employs a small set of anomaly labels to discover significant deviations between abnormal and normal nodes with cross-network meta-learning.
    \item Semi-GNN~\cite{kumagai2021semi} embeds nodes on hyper-sphere space with GCN layer.
    \item ComGA~\cite{luo2022comga} uses community-specific representations with GCN layers to consider local anomalies.
    \item BWGNN~\cite{tang2022rethinking} employs spectral and spatial localized band-pass filters to better handle the ‘right-shift’ phenomenon in detecting anomalies.
    \item ConGNN~\cite{li2024controlled} uses a newly designed diffusion model to augment graphs in AD.
    \item LHML~\cite{guo2022learning} is a few-shot approach to detect nodes that significantly deviate from the majority in learnable hypersphere space.
    \item GHRN~\cite{gao2023addressing} uses homophilic and heterophilic connections with inter-class edge pruning.

\end{itemize}



\spara{Anomaly Detection Results}.
In this experiment, anomaly detection performance of our MGAD is evaluated and compared with three semi-supervised methods and eight unsupervised methods, respectively.
For the evaluation metric, we employ AUC values, which estimates the probability that a randomly selected anomalous node has higher ranking than a normal node.
The anomaly detection results are shown in Table \ref{tab:AUC_results}, and on the basis of the performance results, we present the following observation and analysis:

\begin{itemize}[leftmargin=*]
    \item 
    Among the seven datasets, the proposed MGAD shows the best performance on four datasets, and the second highest on two datasets.
    Specifically, the best result of MGAD in comparison with the baseline methods is 4.91\% higher than second highest method on ACM dataset.
    For the result on Amazon and Pubmed datasets, our MGAD achieves comparable performance. 
    MGAD performs less efficiently in capturing representative features from networks, which contain relatively small number of anomalies and attributes, such as Amazon and Pubmed datasets. 
    Since the context information is derived from attribute features as well as structure information, the small amount of attributes and anomalies contain relatively less context information in Amazon.
    It may be the reason that MGAD performs ineffectively in Amazon.
    CoLA method uses contextual subgraphs in contrastive manner and these techniques seem to work effectively on Pubmed.
    \item 
    When comparing with the shallow deep learning-based methods in unsupervised manner such as AMEN, Radar, and ANOMALOUS and previous early GNN-based methods such as DOMINANT, AnomalyDAE, and ResGCN, our MGAD outperforms these previous approaches.
    The proposed anomaly subgraph makes anomalous node characteristics more distinct, and dual-encoders and multi-stacked GCN layers in the encoder and decoder efficiently extract valuable node features.
    \item %
    For social networks, MGAD shows limited performance.
    This could be largely attributed to the synthetic nature of the anomalies. Structural anomalies involve forming small, densely interconnected cliques by linking unrelated nodes~\cite{ding2019deep,liu2021anomaly,luo2022comga}.
    It is important to note that the selection of 'unrelated nodes' is random, which inherently carries a probability of mistakenly choosing nodes that are, in fact, related. 
    Identifying such inadvertent 'noises' poses a challenge, as they may show similar patterns to normal nodes.
    To provide empirical support for this observation, we conduct an analysis using the Normalized Mutual Information (NMI). 
    NMI is a statistical measure and it compares the similarity between two sets of data, in this case, the anomalies detected by our model versus ComGA. 
    A higher NMI value indicates a greater degree of similarity between the two sets, implying that they find similar anomalies. 
    By analyzing the NMI, which yielded a result of 0.96 for BlogCatalog and 0.94 for Flickr, we infer that both models likely detect similar anomalies. 
    In other words, they might fail to identify anomalies that are inherently more elusive. Based on these findings, we infer that our model is effective in GAD, barring the presence of potential noise elements.
    Based on these findings, we infer that our model is effective in anomaly detection, barring the presence of potential noise elements. 
    The pronounced effect of this phenomenon in social networks could be attributed to inherent characteristics unique to these networks, such as the small world phenomenon~\cite{travers1977experimental}.
\end{itemize}

\begin{figure}[t]
    \centering
    \begin{subfigure}[]{0.49\linewidth}
        \centering
        \includegraphics[width=0.95\linewidth]{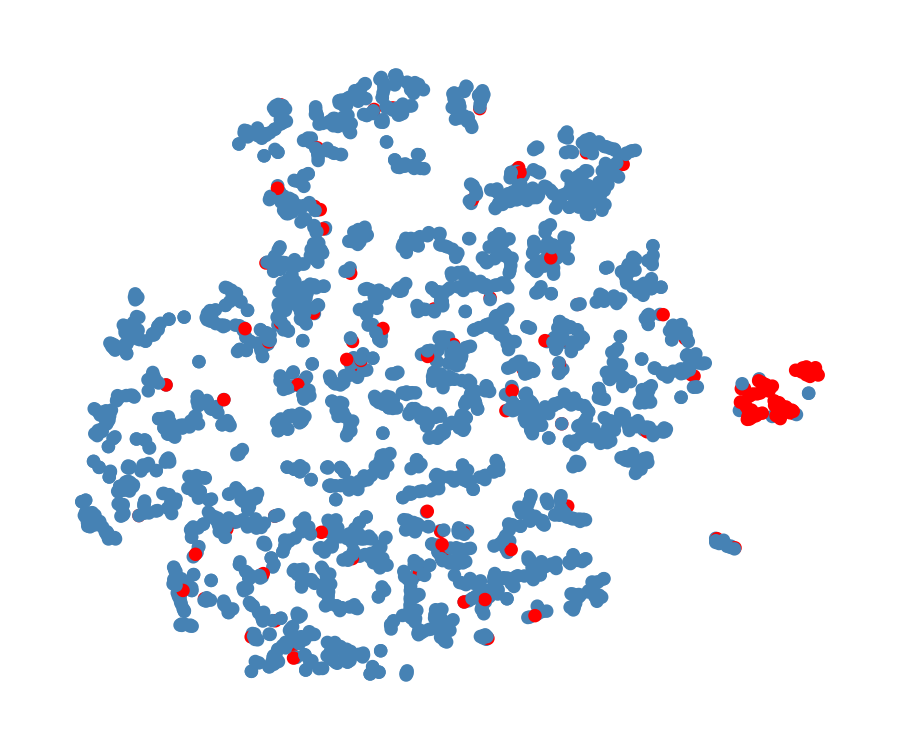}
        \caption{MGAD (ours)}
    \end{subfigure}
    \begin{subfigure}[]{0.49\linewidth}
        \centering
        \includegraphics[width=0.95\linewidth]{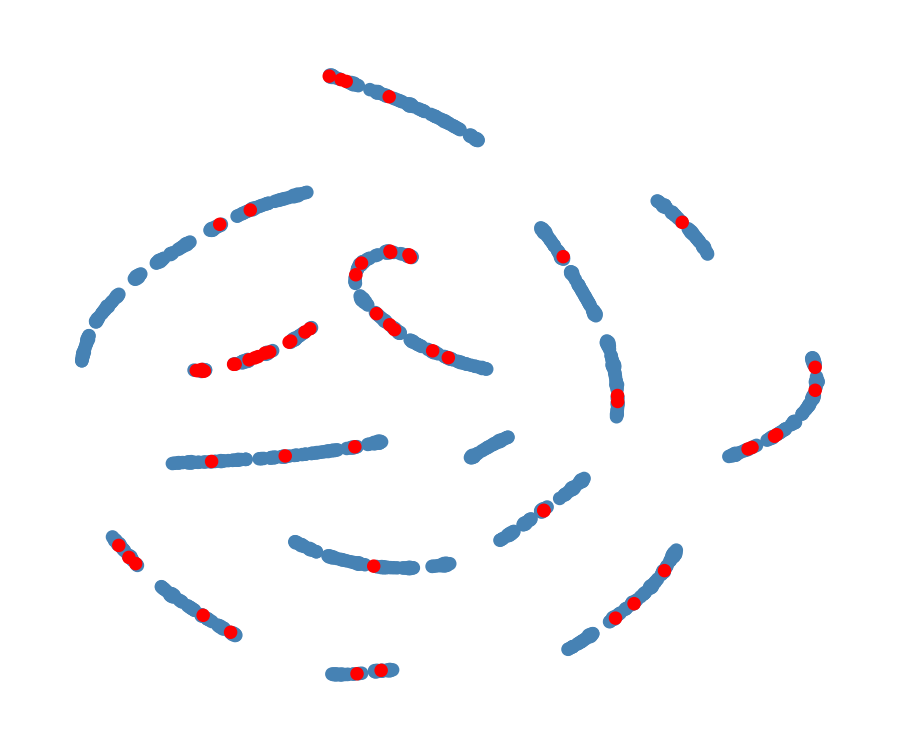}
        \caption{GDN}
    \end{subfigure}
    \begin{subfigure}[]{0.49\linewidth}
        \centering
        \includegraphics[width=0.95\linewidth]{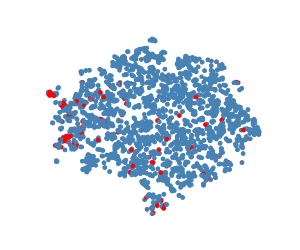}
        \caption{SemiGNN}
    \end{subfigure}
    \begin{subfigure}[]{0.49\linewidth}
        \centering
        \includegraphics[width=0.95\linewidth]{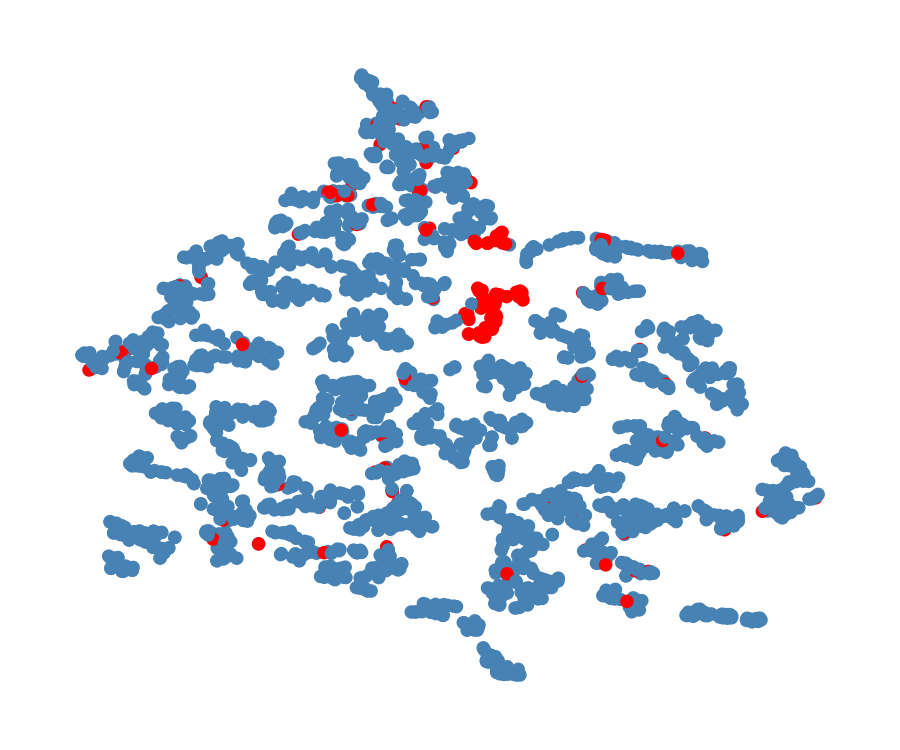}
        \caption{ComGA}
    \end{subfigure}
    \caption{Visualization of node embedding on Cora. The red and blue points represent anomalous and normal nodes, respectively.}
    \label{fig:visualization_cora}
\end{figure}

\begin{table*}[ht]
\centering
\caption{AUC values (\%) on Amazon and average AUC values on seven datasets, respectively.}
\label{tab:real-world_auc}
\resizebox{\textwidth}{!}{%
\addtolength{\tabcolsep}{-3pt} 
\begin{tabular}{@{}lcccccccccccccccc@{}}
\toprule
 & \multicolumn{16}{c}{Amazon} \\ \midrule
 & \multicolumn{1}{l}{AMEN} & \multicolumn{1}{l}{Radar} & \multicolumn{1}{l}{ANOMALOUS} & \multicolumn{1}{l}{DeepSAD} & \multicolumn{1}{l}{CoLA} & \multicolumn{1}{l}{LHML} & \multicolumn{1}{l}{AnomalyDAE} & \multicolumn{1}{l}{Semi-GNN} & \multicolumn{1}{l}{ConGNN} & \multicolumn{1}{l}{ResGCN} & \multicolumn{1}{l}{GDN} & \multicolumn{1}{l}{ComGA} & \multicolumn{1}{l}{DOMINANT} & \multicolumn{1}{l}{BWGNN} & \multicolumn{1}{l}{GHRN} & \multicolumn{1}{l}{MGAD} \\ \midrule
\multicolumn{1}{c}{AUC} & 47.00 & 58.00 & 60.20 & 56.10 & 47.26 & 67.60 & 53.61 & 66.95 & 61.89 & \textbf{71.00} & \underline{68.15} & 63.07 & 62.50 & 58.70 & 51.09 & 65.01 \\ 
\multicolumn{1}{c}{\begin{tabular}[c]{@{}c@{}}Avg.\\ AUC\end{tabular}} & 59.61 & 67.68 & 67.31 & 54.31 & 79.41 & 77.28 & 73.16 & 75.05 & 79.40 & 78.05 & 77.36 & \underline{83.08} & 76.06 & 58.17 & 70.66 & \textbf{85.11} \\ \bottomrule
\end{tabular}%
}
\end{table*}

\spara{Parameter Study}.
In this section, we verify the effectiveness of metapath length $l$ and sampling round $n$.

\begin{itemize}[leftmargin=*]
    \item Effect of metapath length $l$: 
    In this experiment, we investigate effect of metapath length $l$, which is set to 3, 4, and 5.
    The size of anomaly subgraph mainly depends on $l$. When $l$ is larger, the subgraph size is larger too. The performance changes according to each $l$ are depicted in Figure \ref{fig:results_length}.
    We discover a tendency that the smallest anomaly subgraph (length 3) achieves the best performance on most datasets, except for Amazon and BlogCatalog.
    Amazon reaches the highest performance, when the length is 5, the largest subgraph.
    The possible reasons is that the small subgraph on the large graphs may not contain enough patterns between anomaly and normal. On the other hand, the large subgraph on small graphs consists of much unnecessary information since the subgraph and graph become alike.
    %
    As a result, once metapath-based subgraph is used, regardless of its size or length, a novel performance is guaranteed.
    \item Effect of the number of sampling rounds $n$: 
    In this section, we further conduct more experiments in order to prove effectiveness of metapath-based context information between abnormal and normal nodes.
    The number of sampling round $n$ is set to 1, 3, and 5.
    The effect for each $n$ value is shown in Figure \ref{fig:results_sampling_rounds}.
    On most datasets, when value of $n$ is 5, the results show better performance when it comes to max and average AUC values. The one of reasons is that training data is augmented enough according to the number of sampling rounds and through this data augmentation, the subgraph contains more patterns and context information between abnormal and normal nodes for training.
    On extremely dense network, Flickr, the meaningful context might not be extracted for better anomaly detection.
    %
    Consequently, more samplings, more metapath context information, can bring about noticeable performance improvement, since every linking pattern can be sampled around target anomaly nodes and this sampled subgraph can have rich context information.
\end{itemize}

\spara{Sensitivity Analysis}.
In this experiment, we explore sensitivity and robustness of our method according to the ratio of anomalies on five datasets. We conduct this experiments on different ratio of anomalies, such as 1\%, 3\%, 5\%, and 10\% of all labeled anomalies. 
As shown in Figure \ref{fig:results_number_of_anomalies}, performance of MGAD shows its robustness to changes of the number of anomalies used.
Although its performance tends to go higher on ACM and Citeseer, AUC variance on ACM is only 0.1\% and performance become stable with 3\% of anomalies on Citeseer.
In addition, the social network, Flickr, reports decreasing performance as the number of anomaly is larger. 
The possible reason is that Flickr network is much denser than the above-mentioned citation networks.
Since each abnormal node is intricately connected with normal node on the social networks, more information can be extracted from relatively smaller number of anomalies. 
In this reason, employing more anomalies on extremely dense networks become unnecessary information.
Therefore, MGAD performs very stable and robust in spite of using the only small number of anomalies.

\spara{Visualization Analysis}.
In this section, we present visualization analysis about learned node representations in order to evaluate our method.
For visualization, we employ t-distributed stochastic neighbor embeddings~\cite{van2008visualizing} (t-SNE) to reduce the dimensions of the node embeddings from 128 to 2.
Figure \ref{fig:visualization_cora} shows the visualization of node embeddings learned by MGAD, GDN, SemiGNN, and ComGA on Cora.
Our method, MGAD, is capable of discriminating the abnormal and normal node representations efficiently. 
GDN and SemiGNN are not able to effectively separate the abnormal and normal node representations, since its model is not deep enough to exploit the valuable features from the networks.
Consequently, among the four methods, our method, MGAD, shows the most appropriate node representations for graph anomaly detection.

\spara{Ablation Study}.
We conduct the ablation experiments for the subgraph generation module (e.g., 1-hop ego-net). The results of ablation study are as listed in Table~\ref{tab:ablation_subgraph}.
We study the effect of changing communities in the subgraph Generation Module on three datasets (Cora, CiteSeer, Pubmed) and carry out the experiments on three possible scenarios: MGAD without subgraph, MGAD with Ego-network (1-hop), and MGAD with the anomaly subgraph. When MGAD uses the proposed subgraph, it shows the best performance. Specifically, our method effectively augments the few number of label data via label reuse and enriches context information via attention mechanism in generating anomaly communities.

\begin{table}[t]
\centering
\caption{Effect of different communities.}
\begin{tabular}{@{}llll@{}}
\toprule
                   & Cora  & CiteSeer & Pubmed \\ \midrule
MGAD w/o subgraph & 90.21 & 88.53    & 88.82  \\
MGAD w ego-net     & 89.72 & 87.38    & 89.14  \\
MGAD w anomaly subgraph & \textbf{92.29} & \textbf{94.69} & \textbf{92.52} \\ \bottomrule
\end{tabular}
\label{tab:ablation_subgraph}
\end{table}

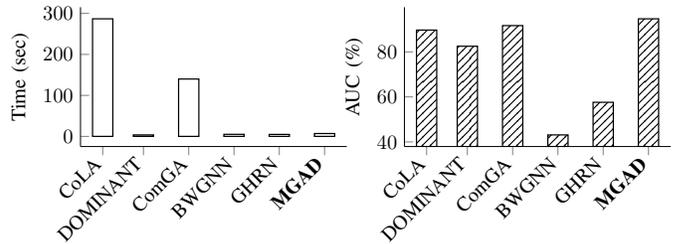
\begin{figure}[]
\centering
\resizebox{\columnwidth}{!}{%
\begin{tikzpicture}
\begin{groupplot}[group style={group size=2 by 1}, height=4cm, width=0.7\columnwidth,
    ybar,
    xtick={1,10,20,30,40,50},
    xticklabels={CoLA,DOMINANT,ComGA,BWGNN,GHRN,\textbf{MGAD}},
    x tick label style={rotate=45,anchor=east},
    axis y line*=left,
    axis x line*=left,
    ylabel near ticks,
    xlabel near ticks
]

\nextgroupplot[ylabel = {Time (sec)}]
\addplot [draw=black, semithick, error bars/.cd, y dir=both, y explicit]
            table[meta=class, x=x, y=y, row sep=\\] {
                x y class \\
                1 286.8 a \\
                10 3.58 a \\
                20 139.9 a \\
                30 5.06 a \\
                40 4.8 a \\
                50 6.99 a \\
            };

\nextgroupplot[ylabel = {AUC (\%)}]
\addplot [draw=black, pattern=north east lines, semithick, error bars/.cd, y dir=both, y explicit]
    table[meta=class, x=x, y=y, row sep=\\] {
                x y class \\
                1 89.68 b \\
                10 82.51 b \\
                20 91.67 b \\
                30 43.11 b \\
                40 57.60 b \\
                50 94.69 b \\
            };
\end{groupplot}
\end{tikzpicture}%
}
\caption{Running efficiency of MGAD and baselines.}
\label{fig:running_efficiency}
\end{figure}

\spara{Running Efficiency}.
We also analyze the running efficiency with other methods on CiteSeer dataset as shown in Figure~\ref{fig:running_efficiency}. In Figure~\ref{fig:running_efficiency} (left), running time of CoLA and ComGA especially reports relatively very long running time, because extracting communities and subgraphs require high computational costs.
Although the other methods record fast running time, these AUC values, Figure~\ref{fig:running_efficiency} (right), are not higher than our MGAD.
Consequently, MGAD shows very efficient running time and the the highest AUC value, while effectively demonstrating the best running efficiency among the baselines.

\spara{Case Study}.
We further examine effecitveness of MGAD on real-world dataset, Amazon, as listed in Table~\ref{tab:real-world_auc}.
These results show that a simple method efficiently detects real-world anomalies and MGAD reports limited performance.
There is no perfect method, which shows the best performance in all scenarios or datasets.
Furthermore, it is an open challenge to efficiently model real-world networks for anomaly detection~\cite{ma2021comprehensive}.
In general, it is non-trivial to achieve novel performance with a fresh idea.
Nonetheless, we successfully adopt metapath, which is used for heterogeneous graph, for homogeneous graph with an idea and improve overall detection accuracy as shown in Table~\ref{tab:real-world_auc} (Avg. AUC).
If the MGAD results are excluded, winner on each dataset is different (e.g., ComGA is winner on ACM), since strong points are quite different from each method.
On the other hand, in overall perspective, MGAD shows the best average AUC and the highest or second highest performance on most datasets, effectively proving its superiority.

\section{CONCLUSION} \label{sec:conclusion}
In this paper, we introduce graph augmentation method to capture connectivity patterns between normal and abnormal nodes in a graph by using both metapath and labels.
We also present a novel metapath-based graph anomaly detection framework, MGAD, which uses GCN layers in both encoder and decoder. 
Specifically, we design metapath-based context information to propagate characteristics between abnormal and normal nodes. 
Moreover, the designed anomaly subgraph effectively augments label information via the number of sampling rounds.
Our model efficiently captures the context information from anomaly subgraphs and whole graph and effectively learns distinct differences in structures and attributes both globally and locally.
The promising results of this study pave the way for future investigations, focusing on the optimization and analysis of metapath patterns.

\bibliographystyle{plain}
\bibliography{references}

\end{document}